\documentclass[11pt]{article}
\usepackage{geometry}                
\usepackage{graphicx}
\usepackage{amssymb}
\usepackage{epstopdf}
\usepackage{xcolor}
\usepackage{soul}
\usepackage[shortlabels]{enumitem}
\usepackage{multirow}
\usepackage{longtable}
\usepackage{subfigure}
\usepackage{amsmath}
\usepackage{lineno}
\usepackage{amssymb}

\newcommand{\bs}{\mathbf{s}} 
\newcommand{\bzero}{\mathbf{0}} 
\newcommand{\bx}{\mathbf{x}} 

\title{Absolute abstraction: a renormalisation group approach}

\author{Carlo Orientale Caputo\thanks{corienta@sissa.it}\\
\small{SISSA - International School for Advanced Studies, 34136 Trieste, Italy}\\
Elias Seiffert\\
\small{University of T\"ubingen, Germany}\\
Enrico Frausin\\
\small{Universit\'a di Trieste, Italy}\\
Matteo Marsili\\
    \small{The Abdus Salam International Centre for Theoretical Physics, 34151 Trieste, Italy}}

\begin{document}

\maketitle

\begin{abstract}
Abstraction is the process of extracting the essential features from raw data while ignoring irrelevant details. 
It is well known that abstraction emerges with depth in neural networks, where deep layers capture abstract characteristics of data by combining lower level features encoded in shallow layers (e.g. edges).
Yet we argue that depth alone is not enough to develop truly abstract representations. 
We advocate that the level of abstraction crucially depends on how broad the training set is. We address the issue within a renormalisation group approach where a representation is expanded to encompass a broader set of data. We take the unique fixed point of this transformation --- the Hierarchical Feature Model -- as a candidate for a representation which is absolutely abstract. This theoretical picture is tested in numerical experiments based on Deep Belief Networks and auto-encoders trained on data of different breadth. These show that representations in neural networks approach the Hierarchical Feature Model as the data get broader and as depth increases, in agreement with theoretical predictions.
\end{abstract}




\bigskip

Abstraction refers to the ability to form general concepts from sensory input or raw data. 
It is well known that representations which are increasingly independent of details arise when raw data is processed in a hierarchy of deeper and deeper layers -- both in-silico~\cite{bengio2017deep} and in-vivo~\cite{marr2010vision}.
Here we argue that universal representations emerge when the process of removing irrelevant details is iterated indefinitely on a universe of data that simultaneously expands in variety, that is when increasing depth is combined with expanding breadth\footnote{We distinguish breadth from width, a term commonly used in the literature to denote the number of variables in different layers.}. These universal representations encode the concept of absolute abstraction.

\begin{figure}[h!]
        \centering
        \includegraphics[width=13cm]{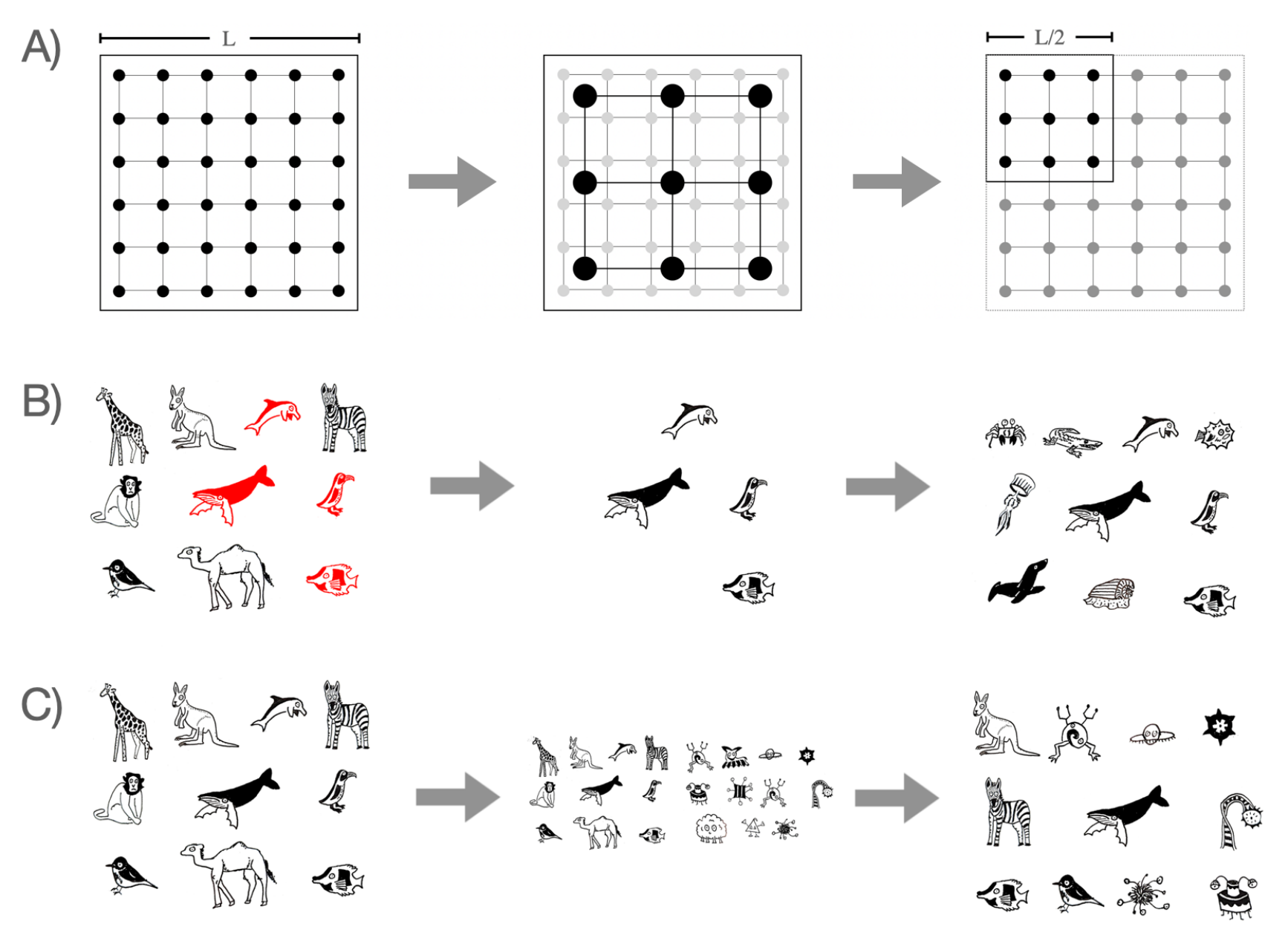}
        \caption{Illustrative example of the RG in statistical physics (A) and of it's application in learning (B and C). The RG (A) entails a coarse gaining (or decimation)  step, whereby low scale degrees of freedom are integrated out, e.g. with the introduction of block variables (large dots in the middle A-panel), and a rescaling step that restores the original size of the system. In a representation of a given domain of items (animals), in B coarse graining is performed zooming into those with a specific feature (living in water) and rescaling corresponds to enriching the representation by adding further details. The same procedure can be reversed (in C): the representation describing a particular domain (animals from planet Earth) is retrained on a wider domain (animals from many planets), neglecting small scale details (e.g. the difference between whales and dolphins).}
        \label{fig:rg}
\end{figure}

We address this in a minimal setting of unsupervised learning from static data (e.g. images). We shall focus on the internal representation\footnote{A representation in this paper is a probability distribution over a set of binary variables.} of the data in a deep layer of a learning machine and on how it adapts when the training data expands in scope. This process can be cast within a renormalisation group (RG) framework that allows us to identify the fixed point of the RG transformation with the 
ultimate outcome of this process when it is repeated indefinitely. The analogy between the process of learning broader and broader data in deeper and deeper layers and the RG in statistical physics~\cite{cardy1996scaling} (see Fig.~\ref{fig:rg} A) is based on the observation that higher order features in learning are akin to large scale properties in statistical models, which are those that the RG singles out. This idea is {\color{black} further} corroborated by similarities between coarse graining and the evolution of representations with depth~\cite{mehta2014exact,koch2020deep}. 

The fact that representations converge to a universal, data-independent, distribution {\color{black} aligns with the Platonic Representation Hypothesis~\cite{huh2024position}, which is receiving more and more empirical support. This states that ``Neural networks, trained with different objectives on different data and modalities, are converging to a shared statistical model of reality in their representation spaces''~\cite{huh2024position}. This hypothesis} 
is consistent with the expectation that a representation encompassing a broader universe of circumstances should be more abstract -- i.e. data independent -- than one describing only a limited domain. In neural networks, this picture suggests that all data-specific information is transferred to the parameters through which the activity of deep layers propagates to the visible layer. In other words, the fixed point of the RG describes how data is organised in the ideal limit of infinite depth and breadth, irrespective of what the data is about. 

Hence this paper approaches the problem of characterising abstract representations on the basis of their sole statistical properties, with no reference to what is being represented, guided only by information-theoretic principles. 

When a representation with a fixed capacity is updated to describe data from a broader domain, low level details need to be sacrificed in order to make space for high level features describing the organisation of the data within the broader domain. Fig.~\ref{fig:rg} C sketches this process for an illustrative example. This process of zooming out to a broader domain while loosing low level details can also be inverted, by zooming into a specific part of the data, thus uncovering low level details (see Fig.~\ref{fig:rg} B). We show that in both cases, the corresponding RG transformation has a unique fixed point which is related to the Hierarchical Feature Model (HFM), recently introduced in~\cite{xie2024simple}. This is reassuring for at least two reasons: First the HFM is a maximum entropy model fully determined by a single sufficient statistics, which is the average level of detail of the features, or the coding cost. This is indeed the only relevant variable in an abstract representation. Second, the HFM satisfies the principle of maximal relevance\footnote{The relevance has been recently introduced~\cite{MMR} as a quantitative measure of ``meaning'' that captures Barlow's intuition~\cite{barlow1989unsupervised} that meaning is carried by redundancy. We refer to Appendix \ref{sec:rel} for a brief discussion of the relevance, or to Ref.~\cite{marsili2022quantifying} for an extended account.}, a principle that has been suggested to characterize most informative representations and that also well-trained learning machines have been shown empirically to obey~\cite{Song,Odilon}. 

The rest of the paper is organised as follows: The next Section places our contribution within the broader literature on machine learning and neuroscience. Then, in Section~\ref{sec:frame} we lay out the framework for our analysis. The RG analysis is presented in Section~\ref{sec:results}. Section~\ref{sec:num} discusses numerical experiments on Deep Belief Networks and auto-encoders that corroborate the theoretical predictions. Within their limited expressive capacity, the networks we studied show that representations in deep layers approach the HFM under the combined effects of depth and breadth. Section~\ref{sec:end} discusses the results and provides some concluding remarks. All technical details are relegated to the Appendix\footnote{The present paper supersedes the preliminary results presented in the Master thesis of one of us~\cite{orientale2023plasticity}.}. 

\section{Literature review}
\label{sec:lit}

Following Marr~\cite{marr2010vision}, we argue that the conceptual underpinnings of the capacity of abstraction are independent of its algorithmic implementation or of whether it is implemented in-silico or in a biological brain. This means that both cognitive neuroscience and machine learning may provide useful insights to characterize abstraction. 

Vision provides the paradigmatic case for exploring how abstract representations arise. Both in biological brains~\cite{marr2010vision} and in artificial neural networks (ANNs)~\cite{bengio2017deep}, vision involves a hierarchical organization of representations: shallow layers detect low-level features, such as edges~\cite{hubel1962receptive}, while deeper layers integrate these features to recognize more abstract, higher-order constructs like objects and faces~\cite{bengio2017deep}. In particular, deeper layers are capable of recognising an object or a face irrespective of it's position, orientation, scale, or of context\footnote{This ability can be promoted in ANN by either augmenting the data using invariances~\cite{simard2002transformation} or by explicitly implementing them in their architecture -- as in convolutional neural networks~\cite{bengio2017deep}. Yet even simple neural networks are able to develop a convolutional structure by themselves~\cite{ingrosso2022data}.}~\cite{dicarlo2007untangling,zoccolan2015invariant}. 
This parallel underscores the broader principle that abstraction emerges through layered processing, regardless of the underlying substrate on which the process is implemented. Yet it has also been argued~\cite{schoenholz2017deep} that the limit of infinite depth is singular, because too much depth without constraints washes away meaningful structure in the input data. 

In general, abstract representations extracted from data have been discussed in terms of ``cognitive maps"~\cite{whittington2022build,tenenbaum2011grow}. 
A cognitive map is not only an efficient and flexible scaffold of data, but it is also endowed with a structure of relations -- uncovered from the data -- that enables abstract computation\footnote{Relational structures such as ``Alice is the daughter of Jim" and "Bob is Alice's brother" allow for computations (e.g. "Jim is Bob's father") which are invariant with respect to the context (Alice, Jim and Bob can be replaced by any triplet of persons that stand in the same relation, in this example)~\cite{whittington2022build}.}~\cite{whittington2022build,tenenbaum2011grow} and supports complex functions. For example, spatial navigation in rats relies on the representation built by several assemblies of specialised neurons, such as grid cells~\cite{o1971hippocampus,rowland2016ten}.

At the highest levels of cognition, representations should integrate data from a broad set of domains, or perceptual modalities~\cite{noppeney2018see}, each of which may be organised according to cognitive maps of a different nature. For example, while visual stimuli are described by object manifolds with supposedly euclidean topology~\cite{dicarlo2012does,ansuini2019intrinsic}, odours have been suggested to be organised in  hyperbolic spaces~\cite{zhou2018hyperbolic}. Higher order representations that integrate the two should therefore be even more abstract, i.e. independent of the data. 

A lot has been understood about the role of depth in learning\footnote{Strictly speaking, most of these insights pertain to supervised learning, yet we assume they reveal properties that are relevant also for unsupervised learning, which is our focus.}~\cite{dicarlo2012does,poggio2017and,cohen2020separability,ansuini2019intrinsic,abbe2021staircase,cheng2024emergence,cagnetta2024deep}. In particular, depth exploits the compositional structure of the data boosting training performance~\cite{poggio2017and,abbe2021staircase,cagnetta2024deep}
and classification capacity~\cite{cohen2020separability}. Indeed, inner layers of ANN portray data that correspond to the same object as ``object manifolds"~\cite{dicarlo2012does} that become better and better separable with depth~\cite{cohen2020separability}, while extracting hierarchies of features that promote taxonomic abstraction\footnote{Taxonomic abstraction is based on the idea that objects are similar if they share the same features (e.g. "cats" and "dogs" both have four legs, a tail, etc) and belong to the same category (mammals). Taxonomic abstraction is fundamentally distinct from thematic abstraction, which is based on co-occurrence between objects that share no features (e.g. "cat" and "sofa"), as discussed in~\cite{mirman2017taxonomic}. In deep neural networks~\cite{abbe2021staircase,cagnetta2024deep} shallow layers capture statistical associations (thematic) while deep layers encode compositional, taxonomic structures. A similar transition is observed in humans with development:
while children favour thematic abstractions, adults more frequently rely on taxonomic (or categorical) structures~\cite{davis2019features}.}. 
It has been proposed that the superior cognitive abilities of {\it homo-sapiens} crucially rely on the expansion of the neo-cortex in depth and to specific mutations that supported reliable neural computation~\cite{namba2024makes}. In this respect, our results suggest that exposure to a broad variety of data and stimuli is also essential for the emergence of abstract representations that may support intelligence. 

The analogy between the processing of data in deeper and deeper layers and the renormalisation group (RG) in statistical physics~\cite{cardy1996scaling} has been suggested long ago~\cite{mehta2014exact,koch2020deep,PhysRevE.97.053304}. This analogy is suggestive because the RG is the theoretical tool to study critical phenomena~\cite{cardy1996scaling} and both artificial and biological learning exhibit critical features~\cite{plenz2021self,MoraBialek,retina,statcrit,xie2022random,sorbaro2019statistical}. Koch {\em et al.}~\cite{koch2020deep} have shown that successive training in deeper layers of neural networks performs an operation similar to coarse graining in the RG. 
Yet the RG transformation does not only involve coarse graining. It also involves rescaling, which is the operation by which the size of the coarse grained system is restored (see Fig.~\ref{fig:rg} A). We argue that this ingredient corresponds to a change in breadth, i.e. in the diversity in the input data or stimuli. It is indeed common sense that a representation encompassing a broader universe of circumstances should be more abstract than one describing only a limited domain. Likewise, we argue that representations in higher levels of the cognitive hierarchy, that integrate a broader set of stimuli, should be more abstract, i.e. independent of the data. 

The emphasis on a data-independent characterisation of abstraction based only of statistical properties contrasts with the ``tuning curve'' approach~\cite{hubel1962receptive,pouget2000information}, in which levels of abstraction are assessed in terms of the features of the data – e.g. edges or faces – that a representation encodes. Likewise, although we implicitly assume that the representations we are interested in describe structured and learnable data, we shall refrain from assuming any specific data structure like the ones proposed in Refs.~\cite{abbe2021staircase,goldt2020modeling,cagnetta2024deep}.


\section{The framework}
\label{sec:frame}

In this paper, a representation is a probability distribution $p(\mathbf{s})$ over the configurations $\mathbf{s}$ of a hidden layer of a generative neural network, such as a Deep Belief Network (DBN)~\cite{hinton2006fast} or an auto-encoders~\cite{hinton2006reducing}. Hence we focus on unsupervised learning and we assume that the network has been trained on some data $\hat{\mathbf{x}}=(\mathbf{x}_1,\ldots,\mathbf{x}_N)$ of $N$ data points, with a non-trivial structure. The representation $p(\mathbf{s})$ is the marginal distribution of activation of variables in the hidden layer induced by the distribution $p(\mathbf{x})$ of the data, i.e. 
\begin{equation}
\label{eq:DBN}
p(\mathbf{s})=\sum_{\mathbf{x}} p(\mathbf{s}|\mathbf{x})p(\mathbf{x})\,.
\end{equation}
We shall confine our discussion to the case where $\mathbf{s}=(s_1,\ldots,s_n)$ is a string of binary variables $s_i=0,1$. Otherwise, we shall keep our discussion as general as possible for the time being. More details on network architectures and the data will be given in Section~\ref{sec:num} and Appendix~\ref{app:DBN}, where we shall test our predictions on specific models. 

We may think of $s_i$ as indicator variables of abstract features: i.e. $s_i=1$ generates points with feature $i$ and $s_i=0$ does not. In order to keep track of indices, when needed, we shall use the notation $\mathbf{s}_{1:n}=(s_1,\ldots,s_n)$ for the state of the hidden layer, and $\mathbf{0}$ (or  $\mathbf{0}_{1:n}$) to indicate the featureless state, i.e. the one with $s_i=0$ for all $i=1,\ldots,n$. 

\section{Renormalization transformations}
\label{sec:results}

The renormalisation group (RG)~\cite{cardy1996scaling} translates the process of focusing on large scale properties of statistical models into a mathematical formalism. As shown in Fig~\ref{fig:rg} A), this process is typically composed of two parts: {\em i)} coarse graining, whereby small scale details are eliminated and {\em ii)} rescaling in order to restore the original (length) scale. The combined effect of these two steps is a transformation from a probability distribution $p(\mathbf{s})$ over the state $\mathbf{s}$ of the system, to a different one $p'=\Re [p]$ whose fixed points $p^*=\Re[p^*]$ describe scale invariant states. Fixed points are endowed with {\em universal} properties which make them independent of small scale details. In statistical physics, $p^*$ depends solely on few fundamental characteristics, such as symmetries and  conservation laws, space dimension and dimensionality of the relevant variables (order parameters)~\cite{cardy1996scaling}. In the context of learning, such universal distributions are natural candidates for an abstract representation, and the RG offers the ideal conceptual framework to search for them.

The representation $p(\mathbf{s})$ of an internal layer of a learning machine depends on its depth but also on the data $\hat{\mathbf{x}}$ used in training. We focus on how a representation changes when it is trained on a larger dataset $\hat{\mathbf{x}}'=(\hat{\mathbf{x}},\hat{\mathbf{x}}_{\rm new})$ incorporating new data $\hat{\mathbf{x}}_{\rm new}$, thus expanding data's domain (Fig.~\ref{fig:rg} C), or when the network is trained only on a subset of the data (Fig.~\ref{fig:rg} B). 

\subsection{Zooming out}
\label{sec:coarse}

We describe how a representation $p(\bs)$ defined in terms of $n$ hierarchical features changes when the data expands to encompass a broader domain. We argue that this transformation entails
\begin{description}
  \item[i)] to account for large-scale features that were not captured by the initial representation
  \item[ii)] to adapt within the limits of finite representational resources and because of this
  \item[iii)] to disregard small-scale details\footnote{It may help to discuss these three steps in a specific example, similar to the one described in Fig.~\ref{fig:rg} C. Imagine how the representation of animals based on observed species in one continent may have been updated when a different continent was discovered. {\bf i)} An expanded universe of observations may lead to the discovery of a new feature that captures the variability in the broader domain. The geographic location may not be the most efficient way to capture this effect, because there may be species found in both continents or traits that are common across them. Hence {\bf ii)} the set of features used to describe the initial dataset may require an update to describe the broader domain. Finally {\bf iii)} fine grained distinctions (e.g. between species of the same genius) of the initial representation need to be disregarded in order to keep the same representational power.}.
\end{description}
The representation capacity is limited by the number $n$ of available features and by the number of bits available to describe a single datapoint, which is measured by the 
entropy $H[\mathbf{s}]=-\sum_{\mathbf{s}}p(\mathbf{s})\log_2p(\mathbf{s})$. Within these limits, we assume that representations maximise a notion of informational efficiency. This means that features are used as parsimoniously as possible, and that new data require the introduction of new large scale features which are in no way related to old ones. 
{This reflects the principle that learning should venture into the unknown with no prejudice.}

As discussed above, we surmise that this transformation takes place under the combined effects of breadth [i)] and depth [iii)], but since we only focus on the representation $p(\mathbf{s})$ we will not need to make this explicit. Section~\ref{sec:num} will explore how breadth and depth affect the internal representations of specific neural networks and whether the predictions of the theory derived here are supported or not in concrete cases.

Formally, let $p(\mathbf{s}_{1:n})$ be a representation over $n$ hierarchical features $\mathbf{s}_{1:n}=(s_1,\ldots,s_n)$ where $s_1$ is the indicator of the largest scale feature while $s_n$ relates to details at the smallest scale. We remind that $s_i$ are binary variables that indicate whether the $i^{\rm th}$ 
feature is present ($s_i=1$) or not ($s_i=0$). 
The transformation $p\to p'=\Re_\uparrow[p]$ discussed above is realised by the following steps: 

\begin{enumerate}
   \item Add a new random feature with $p(s_0=1)=p(s_0=0)=\frac 1 2$ [{\bf i)}], i.e.
\begin{equation}
\label{eq:2cg}
   \tilde p(\mathbf{s}_{0:n})=\frac{1}{2}p(\mathbf{s}_{1:n})
\end{equation}
In this step, the representation is expanded to describe a wider universe of objects. {\color{black} The new feature $s_0$ has a distribution of maximal entropy and it}
is independent of $\mathbf{s}_{1:n}$.
This captures a genuine discovery process characterised by a large scale organisation (described by $s_0$) which {\color{black} is maximally informative and that} cannot be described in terms of combinations of already known features. 
  \item Shift indices $\mathbf{s}_{1:n+1}'=\mathbf{s}_{0:n}$ and renormalise
  \begin{equation}
\label{eq:rescale}
p'(\mathbf{s}_{1:n+1}')=(1-\alpha)\tilde p(\mathbf{s}_{0:n})+\alpha\delta_{\mathbf{s}_{1:n+1}',\mathbf{0}_{1:n+1}}
\end{equation}
where $\alpha$ should be fixed so that the coding cost $H[\mathbf{s}_{1:n}]=H[\mathbf{s}_{1,n}']$ remains the same. This step encodes a reorganisation of features [as in {\bf ii)}] consistent with a principle of parsimony in the use of features. Eq.~(\ref{eq:rescale}) assumes that with this redefinition the featureless state $\mathbf{s}=\mathbf{0}$ -- which corresponds to typical objects that do not require features to be described -- is repopulated in order to respect the constraint on the coding cost $H[\mathbf{s}_{1:n}]$.

  \item Marginalise $s_{n+1}'$ 
  \begin{equation}
\label{ }
p'(\mathbf{s}_{1:n}')=\sum_{s_{n+1}=0,1}p'(\mathbf{s}_{1:n+1})\,.
\end{equation}
In this step, the most detailed feature is eliminated, analogously to what happens in the coarse graining step of the RG [{\bf iii)}].
\end{enumerate}

Without the second step, i.e. with $\alpha=0$, this RG procedure would converge to a distribution $p(\mathbf{s})=2^{-n}$ of independent variables. This, as we shall see, is a possible (although trivial) fixed point with $H[s_{1:n}]=n$ (in bits). When mixing $\tilde p$ with a deterministic term $\delta_{\mathbf{s}_{1:n+1}',\mathbf{0}_{1:n+1}}$, the coding cost $H[\bs]$ decreases whereas in the other two steps it increases, because the $n^{\rm th}$ feature is replaced with a totally random one. Hence  there is a unique solution for $\alpha$ (see Appendix~\ref{app:alpha} for more details). 

\begin{figure}[h!]
        \centering
        \includegraphics[width=8cm]{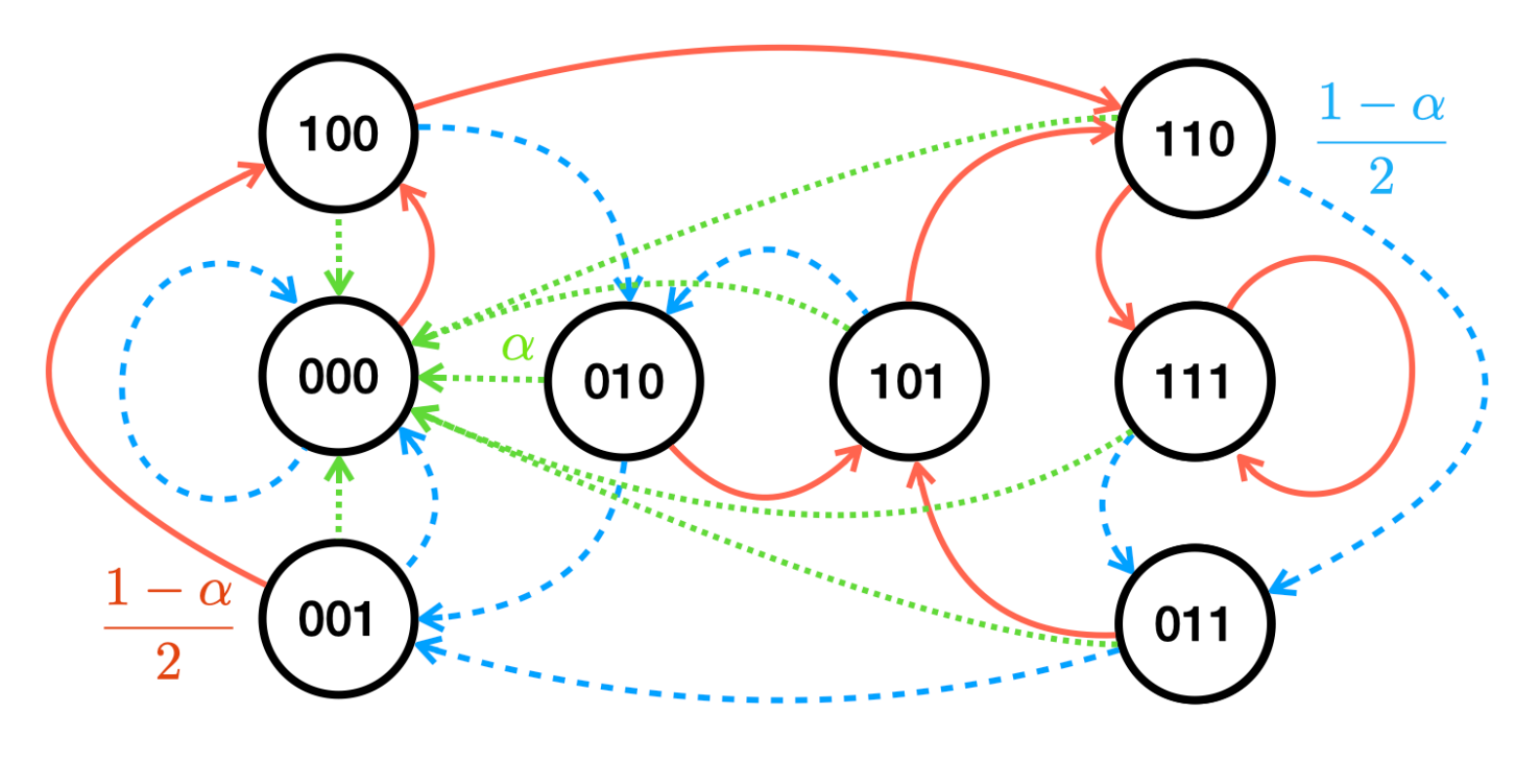}
        \caption{Graphical representation of the transition matrix $T_{\mathbf{s}_{1:n},\mathbf{s}_{1:n}'}$ of the coarse graining RG for $n=3$. States are represented by circles and transitions by arrows. From each state $\mathbf{s}_{1:n}$ there are only three non-zero matrix elements of $T_{\mathbf{s}_{1:n},\mathbf{s}_{1:n}'}$. Two of them correspond to the possible states $\mathbf{s}_{1:n}'=(s_0,\bs_{1:n-1})$ that can be reached by either adding $s_0=1$ to the left of $\bs_{1:n}$ (red links) or adding $s_0=0$ (blue dashed links). Both transitions occur with probability $T_{\mathbf{s}_{1:n},\mathbf{s}_{1:n}'}=\frac{1-\alpha}{2}$. The third transition resets to the $\mathbf{0}_{1:n}$ state (green dotted links) and it occurs with probability $T_{\mathbf{s}_{1:n},\mathbf{0}_{1:n}}=\alpha$.}
        \label{fig:debruijn}
\end{figure}

A simple argument shows that the transformation $p\to p'=\Re_\uparrow[p]$ converges to a unique fixed point. This is because there is a monotonous relation between $H[\bs]$ and $\alpha$. For a fixed $\alpha$, the transformation described above is linear, which means that it can be expressed as
\begin{equation}
\label{eq:Markov}
p'(\mathbf{s}_{1:n}')=\sum_{\mathbf{s}_{1:n}}p(\mathbf{s}_{1:n})T_{\mathbf{s}_{1:n},\mathbf{s}_{1:n}'}
\end{equation}
where $\hat T$ is a stochastic matrix. The associated Markov chain describes a random walk with resetting~\cite{Evans_2020} on the de Bruijn graph~\cite{de1946combinatorial}, and it is shown in Fig.~\ref{fig:debruijn} for $n=3$. 
This Markov chain is clearly ergodic, because $\hat T^m$ has all strictly positive elements for $m\ge n$. This is because each time the variable $s_0$ is generated at random, so after $m$ iterations, every state $\mathbf{s}_{1:n}'$ can be generated. By the theory of Markov chains\footnote{We recall the Perron-Frobenius theorem, which states that a matrix with all positive elements has a unique maximal eigenvalue whose corresponding eigenvector has all positive elements. For an ergodic stochastic matrix this eigenvalue is one and the (left) eigenvector is $p^*$ in our case.}, under successive applications of the transformation, the distribution converges to a fixed point $p^*$ from any initial distribution $p$, and the limit is unique. The same necessarily applies to the transformation where $H[\bs]$ is held fixed.

The unique fixed point $p^*$ of the RG transformation, in terms of the parameter $\alpha\in [0,1]$, is given by (see Appendix~\ref{app:fixedpoint})
\begin{equation}
p^*(\bs)=\frac{2\alpha}{1+\alpha}\left(\frac{1-\alpha}{2}\right)^{m_{\bs}}+\frac{1-\alpha}{1+\alpha}\left(\frac{1-\alpha}{2}\right)^{n}
\label{fixed_point_out}
\end{equation}
where 
\begin{equation}
\label{eq:defms}
m_{\bs}=\max\{i:~s_i=1\},~~~~(\bs\neq\mathbf{0})
\end{equation}
is the index of the most detailed active feature in $\bs$, {\color{black} that we shall call the {\it level of detail}}, and $m_{\mathbf{0}}=0$. Interestingly, this distribution is related to the Hierarchical Feature model (HFM), introduced recently in Ref.~\cite{xie2024simple}, which is defined as
\begin{equation}
\label{eq:HFM}
h_n(\bs)=\frac{1}{Z_n}e^{-g m_{\bs}},\qquad Z_n=1+\frac{1-(2e^{-g})^n}{e^g-2}\,,
\end{equation}
where $g\ge 0$ is a parameter {\color{black} (see Appendix~\ref{sec:HFM} for a discussion of the HFM)}. 

Indeed, with $\alpha=1-2e^{-g}$ one can show (see Appendix~\ref{app:fixedpoint}) that $p^*$ coincides with 
the marginal distribution of the $n$ most coarse grained features of an HFM with infinite features:
\begin{equation}
\label{eq:pstar}
p^*(s_{1:n})=\lim_{m\to\infty}\sum_{s_{n+1:m}}h_m(s_{1:m})\,.
\end{equation}

We shall come back to the significance of this result after discussing an analogous transformation that proceeds in the opposite direction,  describing how a representation changes when zooming into a part of the dataset.

\subsection{Zooming in}

The inverse transformation $\Re_\downarrow$ to the one described above, describes how the internal representation of a learning machine changes, in ideal circumstances, when we focus on a specific subclass of objects while enriching the data of further small scale details. Again we shall require that the coding cost $H[\bs]$ of the representation remains constant in this process. The same general arguments as those discussed above for $\Re_\uparrow$ apply.
The fine graining transformation $p\to p'=\Re_\downarrow[p]$  is based on the following steps. 
\begin{enumerate}
  \item Zooming in on objects with $s_1=1$
  \begin{equation}
\label{ }
p(\mathbf{s}_{2:n})=p(\mathbf{s}_{2:n}|s_1=1)
\end{equation}
 \item Shift indices $\mathbf{s}_{1:n-1}'=\mathbf{s}_{2:n}$ and define $\tilde p(\mathbf{s}_{1:n-1}')=p(\mathbf{s}_{2:n})$.
 \item Add a new feature $s_n'$ 
 \begin{equation}
\label{ } 
p'(\mathbf{s}_{1:n}')=p'(\mathbf{s}_{1:n-1}'|s_n')p(s_n')
\end{equation}
where $p(s_n'=1)=q=1-p(s_n'=0)$ is determined by requiring that the new representation has the same coding cost as the old one, i.e. $H[\mathbf{s}_{1:n}']=H[\mathbf{s}_{1:n}]$.
Also we assume
\begin{eqnarray}
p'(\mathbf{s}_{1:n-1}' |s_n'=0) & = & \tilde p(\mathbf{s}_{1:n-1}')\label{maxign1} \\
p'(\mathbf{s}_{1:n-1}' |s_n'=1) & = & 2^{-n+1}\,.\label{maxign}
\end{eqnarray}
    The first of these two equations implies that the representation without the new feature is the same as the original representation over $\mathbf{s}_{2:n}$. The second equation enforces a maximum ignorance principle whereby the presence of the $n^{\rm th}$ feature ($s_n=1$) does not provide any information on whether more coarse grained features are present or not. This is the equivalent of the first step of the procedure of Sect.~\ref{sec:coarse}, where higher order features are introduced independently of lower order ones. Indeed, by Eq.~(\ref{eq:2cg}), $p'(s_1'|\bs_{2:n}'\neq \mathbf{0}_{2:n})=p(s_1')=\frac 1 2$.
\end{enumerate}
We can appeal to the same arguments as in Section~\ref{sec:coarse} to show that the transformation $p\to p'=\Re_\downarrow[p]$ has a unique fixed point\footnote{The transformation $p\to p'=\Re_\downarrow[p]$ is given by $p'(\bs_{1:n}')=\frac{q}{2^n}\delta_{s_n',1}+(1-q)p(\bs_{2:n}|s_1=1)\delta_{s_n',0}$. Strictly speaking, this is not a linear transformation because of the conditional probability $p(\bs_{2:n}|s_1=1)$ that appears in the right hand side. Yet it satisfies all the conditions of the non-linear Peron-Frobenius theorem~\cite{lemmens2012nonlinear} that leads to the same conclusions of the existence of a unique fixed point $p^*=\Re_\downarrow[p^*]$. In particular any state $\bs'$ can be reached with finite probability from any other state $\bs$ in a finite number of steps (irreducibility and primitivity).}. 
It is also easy to check that the fixed point is the HFM, i.e. 
\begin{equation}
p^*(\bs_{1:n})=h_n(\bs_{1:n})\,,
\label{pstarin}
\end{equation}
with a value of $g$ that depends on $H[\bs]$ (or $q$). Indeed if Eq.~(\ref{pstarin}) holds, then Eq.~(\ref{maxign1}) is satisfied because $p^*(\bs_{2:n}|s_1=1)=h_{n-1}(\bs_{2:n})=p^*(\bs_{1,n-1}|s_n=0)$ and the HFM satisfies the condition~(\ref{maxign}) by definition.

\subsection{Significance of the results}

The fact that the fixed point of the RG transformation is related to the HFM is interesting. Let us discuss the different aspects:

\paragraph*{Data-independence of the internal representation} If the internal representation $p(\bs)$ approaches a universal distribution with depth and breadth, then $p(\bs)$ does not contain any information on the specific nature of the data that it represents. Data specific information is stored in the conditional distribution $p(\mathbf{x}|\mathbf{s})$ that generates data points $\bx$ from a given internal state $\bs$. The internal representation $p(\mathbf{s})$ captures merely the way in which data is organised into combinations of features that are reproduced by the parameters learned by each layer combining the features learned by the earlier layer. Ref.~\cite{xie2024simple} argues that an architecture where $p(\bs)$ is held fixed provides advantages which make learning more similar to understanding.

\paragraph*{Level of detail as sufficient statistics} 
The HFM was introduced in~\cite{xie2024simple} as an efficient scaffold for data based on two principles. First, the HFM assumes that features are organised in a hierarchical scale of detail and it requires that the occurrence of a feature $s_k=1$ at level $k$ does not provide any information on whether lower order features are present or not. This means that, conditional on $s_k=1$, all lower order features are as random as possible, i.e. $H[\mathbf{s}_{1:k-1}|s_k=1]=k-1$ in bits. This requirement implies that the distribution $p({\mathbf{s}})$ should be a function of $m_{\mathbf{s}}$, as defined in Eq.~(\ref{eq:defms}), {\color{black} i.e. of the level of detail of objects associated to state $\bs$.} This is equivalent to requiring that the level of detail $m_{\bs}$ is the only sufficient statistics of the distribution, in the sense that knowledge of the probability $p(m_{\bs}=k)$ is sufficient to specify the full distribution $p({\mathbf{s}})$. If one further requires that just the average level of detail $\bar m=\langle m_{\bs}\rangle$ is sufficient to determine $p({\mathbf{s}})$, i.e. that $p({\mathbf{s}})$ is a maximum entropy distribution, then this implies that the dependence of $p(\bs)$ on $m_{\bs}$ should take an exponential form, which singles out Eq.~(\ref{eq:HFM}) as the only possible solution. 

\paragraph*{Relation with the principle of maximal relevance} 
The maximum entropy requirement also coincides with demanding that $p(\bs)$ satisfies the principle of maximal relevance. The relevance is defined~\cite{marsili2022quantifying} as the entropy $H[E]$ of the coding cost $E_{\bs}=-\log p(\bs)$ and is discussed in further detail in Appendix~\ref{sec:rel}. For our discussion, let it suffice to say that the relevance is a model-free  measure of ``meaningful'' information content in a representation or in a data-set. Indeed, the internal representation of learning machines trained on non-trivial real data approach closely distributions of maximal relevance~\cite{Song,Odilon,marsili2022quantifying}. The principle of maximal relevance predicts that the number of states $\bs$ with coding cost equal to $E$ should increase exponentially in $E$, i.e. $W(E)=W_0 e^{\nu E}$~\cite{statcrit,marsili2022quantifying}. The HFM satisfies this property with $\nu=\log 2$ because the number of states with $m_{\bs}=k$ equals $2^{k-1}$ ($k>0$). 
Linearity of $E_{\bs}$ with $m_{\bs}$ implies that the number of bits required to describe a state $\bs$ increases linearly with its level of detail. 

\paragraph*{Statistical mechanics of the HFM and the thermodynamics of thought process} 
In the limit $n\to\infty$ the HFM features a phase transition at $g_c=\log 2$ between a random phase where the entropy $H[\mathbf{s}]$ and the average number of features $\langle m_{\bs}\rangle$ are of order $n$ for $g<g_c$, and a ``low temperature'' phase where $h_n(\mathbf{s})$ is dominated by a finite number of states, with $H[\mathbf{s}]$ and $\langle m_{\bs}\rangle$ attaining finite limits as $n\to\infty$. The distribution $h_n(\bs)$ for $g=g_c$ reproduces Zipf's law, a statistical regularity that characterises many efficient representations, from language~\cite{balasubrahmanyan2002algorithmic}, to assemblies of neurons~\cite{retina} and the immune system~\cite{Immune}.
We refer to~\cite{xie2024simple} for further discussion of the properties of the HFM. 

The HFM is particularly intriguing for the peculiarity of its free energy landscape. The fact that the number of states $W(E)$ at energy $E$ grows exponentially with $E$ implies that the entropy $S(E)=k_B\log W(E)$ is linear in $E$. This means that also the free energy $F(E)=E-TS(E)$ is linear in $E$ and that there is a temperature for which $F(E)=F_0$ is constant. At this temperature, transitions from states of different energy require no work, in principle. 
These transitions, where either more details are added (increasing the level of detail $E$) or removed (decreasing $E$), are the natural building blocks of thought processes like generalization or specialization. 

\paragraph*{Marginalization properties of the HFM}

For $g\le g_c$ the constant $\alpha=1-2e^{-g}$ in the distribution $p^*$ of Eq.~(\ref{fixed_point_out}) becomes negative. So the marginal HFM Eq.~(\ref{eq:pstar}) describes the fixed point of the transformation where the universe of the data expands only for $g\ge g_c$. Indeed, marginalising over the high order features yields a mixture between the HFM and the uniform (maximum entropy) distribution
(see Appendix~\ref{app:fixedpoint}) 
\begin{equation}
\label{eq:margtop}
\sum_{{\bs}_{n+1:m}}h_m({\bs}_{1:m})=\frac{Z_n}{Z_m} h_{n}(\mathbf{s}_{1:n})+\left(1-\frac{Z_n}{Z_m} \right) 2^{-n}\,.
\end{equation}
Eq.~(\ref{fixed_point_out}) coincides with the limt $m\to\infty$ of this expression, which returns the uniform distribution $p(\bs)=2^{-n}$ when $g\le g_c$. This corresponds to the degenerate limit when $H[\mathbf{s}]= n$ (in bits). For all values $H[\mathbf{s}]< n$ the fixed point has $g>g_c$. 

On the other hand, marginalising $h_n(\mathbf{s}_{1:n})$ over the low order features $\mathbf{s}_{1:k}$ returns a mixture between a HFM over the remaining $n-k$ features and the featureless state
\begin{equation}
\sum_{\mathbf{s}_{1:k}}h_n(\mathbf{s}_{1:n})=\frac{\xi^k Z_{n-k}}{Z_n}h_{n-k}(\mathbf{s}_{k+1:n}) + \frac{(1-\xi/2)(\xi^k-1)}{(\xi-1)Z_n}\delta_{\mathbf{s}_{k+1:n},\mathbf{0}_{k+1:n}}
\,,
\end{equation}
where $\xi=2e^{-g}$. Marginalising over an infinite number of low order features yields
\begin{equation}
\label{ }
\lim_{k\to\infty} \sum_{{\bs}_{-k+1:n}}h_{n+k}({\bs}_{-k+1:n})=
\left\{\begin{array}{ll}
\delta_{\bs_{1:n},{\bzero}_{1:n}} & \hbox{if $g\ge g_c$} \\
\left(1-a\right) h_n(\bs_{1:n})+ a \delta_{\bs_{1:n},{\bzero}_{1:n}} & \hbox{if $g<g_c$} \end{array}\right.
\end{equation}
with $a=(2\xi-1)\xi^{-n}$. Therefore integrating out low or high order features leads to degenerate distributions -- either the one concentrated on the featureless state or the totally random one. This is consistent with the fact that coarse graining alone is not sufficient to define a RG. New information has to be injected at each RG step. 

\section{Empirical evidence in Deep Neural Networks}
\label{sec:num}

In this Section we test the ideas discussed in previous Sections on two architectures, Deep Belief Networks (DBNs) and auto-encoders (AE), trained on different datasets which are variants of the MNIST dataset of handwritten digits. A full account of architectures, algorithms and datasets is given in Appendix~\ref{app:numerics}. We retain the essential details in what follows. 

\subsection{Comparing internal representations with the HFM}
\label{sec:gauge}

As a measure of the distance of representations to the HFM we take the Kullback-Leibler (KL) divergence between the empirical distribution of internal layers and the HFM. The empirical distribution is obtained either as the distribution of clamped states, i.e. of states obtained propagating each datapoint through the layers of  the network, or sampling the distribution by Montecarlo methods. 

We first observe that the $n$ hidden binary variables $\bs$ of the internal representation may not coincide with the variables $\bs'$ that appear in the HFM. Thus it is necessary to define the transformation $\bs\to\bs'$. 

First note that the two values $s_i=0$ or $1$ of each variable can be associated in two different ways to the values of $s_i'$. Hence a state $\bs$ defined by $n$ hidden binary variables, admits $2^n$ possible states $\bs'$, each consistent with the ``gauge'' $\mathbf{\tau}$ that defines the transformation $s_{i}\to s_i' =\tau_i s_{i}+(1-\tau_i)(1-s_{i})$ with $\tau_i=0,1$. In order to fix this gauge we set $\mathbf{\tau}$ such that the most frequently sampled state in each representation corresponds to the featureless state $\mathbf{s}'=\mathbf{0}$, as for the HFM. In addition, while in the neural networks the variables $s_i$ are {\it a priori} equivalent, in the HFM they are not, because they are hierarchically organised. There are $n!$ possible ways to order the variables $\bs_{1:n}'$, that correspond to a different HFM. 
For a given permutation $\pi=(\pi(1),\ldots,\pi(n))$ of the integers $1,\ldots,n$, and a given gauge $\mathbf{\tau}$, the combined effect of these two operations defines the transformation
\begin{equation}
\label{ }
\mathbf{s}'=\mathcal{G}_{\mathbf{\tau},\pi}(\mathbf{s}),\qquad s_i' =\tau_{\pi(i)} s_{\pi(i)}+(1-\tau_{\pi(i)})(1-s_{\pi(i)})\,.
\end{equation}
The transformation $\bs\to\bs'$ that is used to compare internal representations of neural networks with the HFM is obtained minimizing their KL divergence
\begin{equation}
\label{eq:defs1}
    \bs\to\bs'={\rm arg}\min_{\pi,g}\,D_{KL}\left[\hat p(\bs)|\!|h_n\left(\mathcal{G}_{\mathbf{\tau},\pi}(\mathbf{s})\right)\right]\,,
\end{equation}
where the minimum is taken over all permutations and on the parameter $g$ of the HFM. In practice, the minimum over permutations is carried out by a greedy heuristics that compares two permutations that differ by the swap of two indices. We consider all possible swaps of indices recursively until no improvement is possible.
All results of this Section are derived performing this transformation. 

\subsection{Deep Belief Networks}

Deep Belief Networks (DBNs) are layered networks composed of stacked Restricted Boltzmann Machines (RBMs), whereby the hidden layer of one RBM coincides with the visible layer of the deeper RBM. The $\ell^{\rm th}$ hidden layer, which is the hidden layer of the $\ell^{\rm th}$ RBM, contains $n_\ell$ binary variables $\bs_\ell\in\{0,1\}^{n_\ell}$. 

We trained DBNs on datasets of increasing breadth by successively training the RBMs that connect its layers~\cite{hinton2006fast}. Starting from the data $\hat{\mathbf{x}}\equiv\hat{\mathbf{s}}^{(0)}$, we train layer $\ell=1,2,\ldots,L$ from the dataset $\hat{\mathbf{s}}^{(\ell -a)}$ by maximising the likelihood 
\begin{equation}
\label{ }
\mathcal{L}_\ell(\theta_\ell)=\sum_{\bs_{\ell-1}}\hat p(\bs_{\ell-1})\log \sum_{\bs_{\ell}}p(\bs_{\ell-1},\bs_{\ell}|\theta_\ell)
\end{equation}
over the parameters $\theta_\ell$ of the joint distribution $p(\bs_{\ell-1},\bs_{\ell}|\theta_\ell)$. The architecture used is the same as that used in Ref.~\cite{Song} with ten layers of width $n_\ell=500,250,120,60,30,25,20,25,10$ and $5$ for $\ell=1,2,\ldots,10$. For each trained network we make sure that it correctly generates data-points that are statistically similar to those in the dataset, avoiding known pathologies of these networks~\cite{hinton2006fast}. We discuss more details on the architecture, datasets and training algorithms in Appendix~\ref{app:DBN}. Here we focus on the results.

\begin{figure}[h!]
        \centering
        \includegraphics[width=15cm]{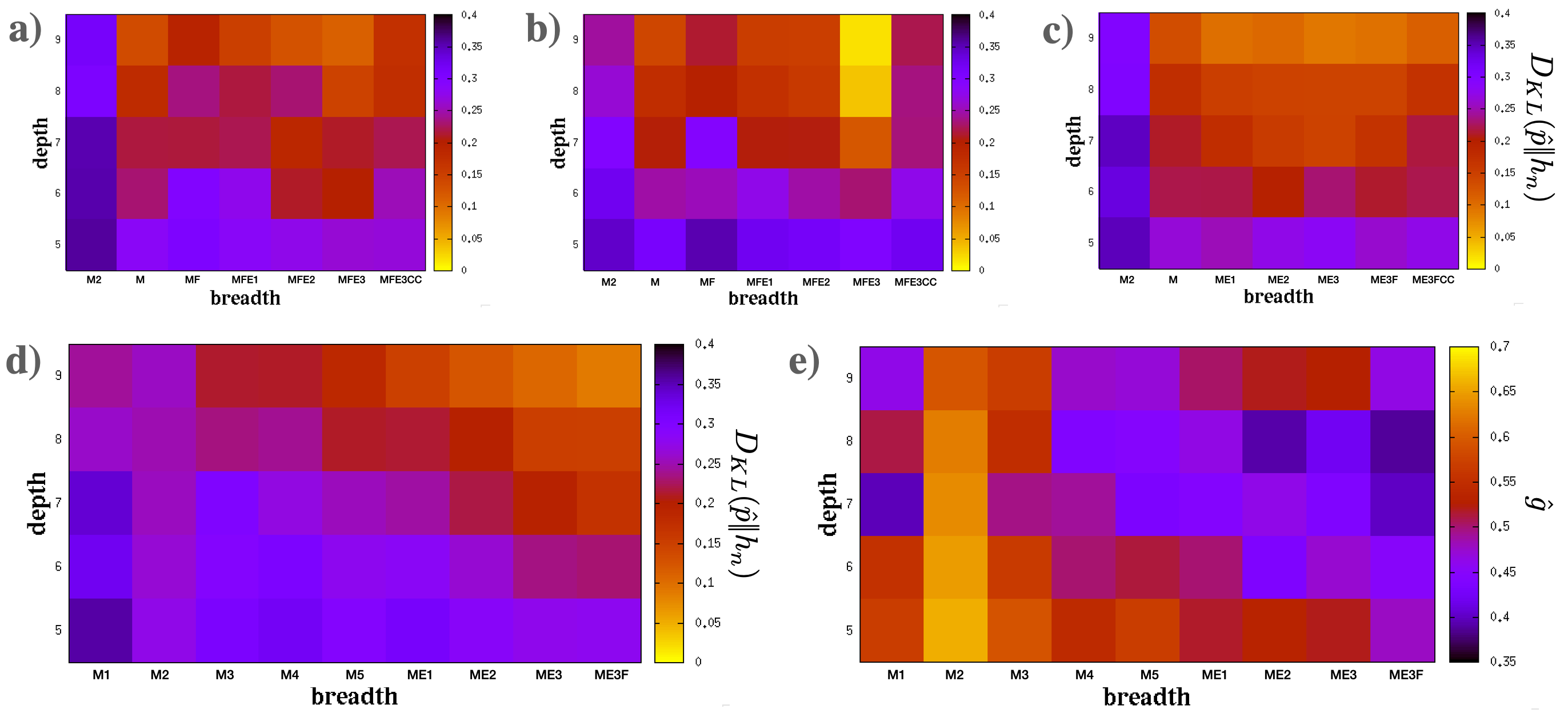}
        \caption{Color plot of the $D_{KL}$ divergence per node between the internal representation of a layer inside a DBN and the HFM ({\bf a, b, c} and {\bf d}) and of the fitted parameter $g$ ({\bf e})for the DBN in panel {\bf d)}. The internal representation is obtained sampling $10^5$ configurations $\bs$ for each layer from the equilibrium distribution. The DBN's are trained on datasets of increasing breadth: in {\bf  a), b)} and {\bf c)}  M2 refers to a dataset obtained by transforming the digits $2$ of MNIST with translations and rotations, to obtain $N=6\cdot 10^4$ data points, and M refers to the complete MNIST dataset. In {\bf  a)} and {\bf b)} MF refers to MNIST plus Fashion MNIST, MFE1 coincides with MF plus the letters from $a$ to $i$ of EMNIST, adding letters up to $r$ generates MFE2 and adding all letters yields MFE3. MFE3CC contains all the data of MFE3 plus Cifar-10 images, rescaled to $28\times 28$ pixels. The DBNs in {\bf a)} and {\bf b)} are trained for $10^5$ and $10^4$ epochs respectively. DBNs in {\bf c)} and {\bf d)} are also trained for $10^5$ epochs.
        In {\bf c)} the order with which the data sets are learned is changed: After MNIST the DBN is trained  first with letters in EMNIST and then with images of Fashion MNIST and finally Cifar-10 is added (ME3FCC). 
        In {\bf d)} the datasets are fragmented in different ways. First MNIST is divided into 5 parts (M1 with digits 0 and 1, M2 up to 3, M3 up to 5, M4 up to 7 and M5 with all digits) then EMNIST and Fashion MNIST are added as before. For this network, the fitted value of the parameter $g$ of the HFM is shown in panel {\bf e)}.}
        \label{fig:DKL_DBN}
\end{figure}

The results of extensive numerical experiments on DBNs are presented in Fig.~\ref{fig:DKL_DBN}, which displays the Kullback-Leibler divergence per node\footnote{In the DBN we studied the number of hidden nodes $n_\ell$ vary substantially with depth $\ell$. We expect $D_{KL}(\hat p_\ell|\!|h_{n_\ell})$ to be proportional to $n_\ell$, which is why we show the Kullback-Leibler divergence per node.} $D_{KL}(\hat p_\ell|\!|h_{n_\ell})/n_\ell$ in colour code (see bar on the left) as a function of breadth and depth (for layers\footnote{We could not reliably estimate the distribution $p(\bs)$ for shallower layers.} $\ell=5$ to $9$). 
We found that $D_{KL}(\hat p_\ell|\!|h_{n_\ell})/n_\ell$ generally decreases when both depth and breadth increase, as suggested by the RG approach. For a fixed depth, the internal representation initially approaches the HFM as breadth increases and subsequently diverges from it. This suggests that both depth and breadth are necessary for abstraction to emerge, as claimed in the theoretical analysis of the previous Section. 

{\color{black} Fig.~\ref{fig:DKL_DBN} suggests that this conclusion is robust with respect to varying the training time, the order in which datasets are learned and how they are segmented. Specifically,} panels {\bf a)} and {\bf b)} compare the same DBN trained for $10^5$ ({\bf a}) or $10^4$ ({\bf b}) epochs. {\color{black} We found evidence that convergence to stable distributions requires long training times, as observed by Decelle and co-workers~\cite{decelle2021equilibrium}. Consistently, we found that, although the approach to the HFM is observed in both cases, the convergence is smoother for longer training times ({\bf a}) than for shorter times ({\bf b})\footnote{Increasing further the training time (to $10^6$ epochs) led to the collapse of representations to distributions sharply peaked on one or two states.}.} Panels {\bf a)} and {\bf c)} compare the same DBNs trained (for $10^5$ epochs) on the same datasets learned in a different order: in {\bf a)} Fashion MNIST is introduced before EMNIST while in {\bf c)} we did the opposite. This comparison shows that{\color{black}, although both cases qualitatively agree with our thesis,} the order in which datasets are learned matters: adding datasets which are more similar to those already learned results in a smoother approach to the HFM. The comparison between panels {\bf c)} and {\bf d)} -- where the MNIST dataset is disaggregated in 5 parts which are added sequentially -- further corroborates this conclusion\footnote{In the DBNs in {\bf d)} the addition of the Cifar-10 dataset to ME3F (see caption) led to the collapse of the representation (which is why it was not shown). This is a manifestation of the limits of the representation capacity of these neural networks~\cite{montufar2016restricted}.}. Panel {\bf e)} shows that the fitted values of the parameter $g$ of the HFM decrease with breadth. This is consistent with the fact that in order to represent a wider universe of data, the internal representation needs to expand (indeed the entropy $H[\bs]$ is a decreasing function of $g$). 
Results obtained fitting the marginal HFM (Eq.~\ref{eq:pstar}) {\color{black} lead to similar conclusions}.

\subsection{Auto-encoders}
\label{sec:AE}

An auto-encoder (AE) is a neural network based on two main components: an encoder, which maps input data to a lower-dimensional latent space, and a decoder, which reconstructs the original data from this latent representation. Both the encoder and the decoder are neural networks of $L\ge 1$ layers, which are jointly trained to minimize reconstruction error. 
We trained AE of different depth $L$ on different datasets, keeping a fixed number $n$ of variables in the latent representation. 
We refer to Appendix~\ref{app:AE} for further details on the architectures used. 

\begin{figure}[h!]
        \centering
        \includegraphics[width=14cm]{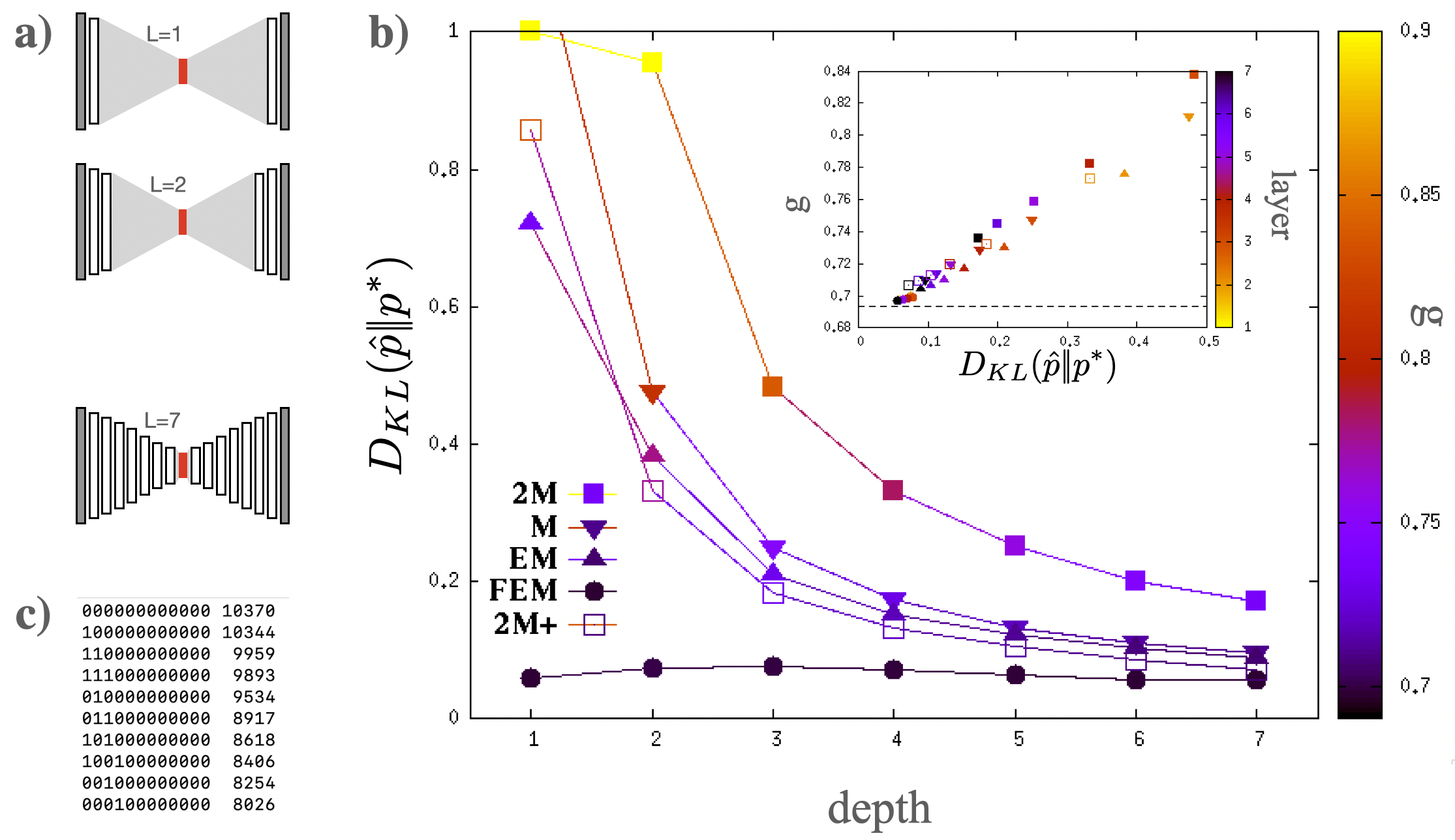}
        \caption{{\bf a)} Architectures of the AE used in this study. The hidden layers (white sticks) process information from the visible layer of the encoder (in grey, left) to the bottleneck layer (in red), which is then propagated through a specular sequence of hidden layers to the visible layer of the decoder (in grey). The architecture with $L+1$ layers is obtained adding an extra hidden layer between the last one of the AE with $L$ layers, on both sides of the bottleneck. {\bf b)}
        Kullback-Leibler divergence from the marginal HFM $p^*$ of Eq.~(\ref{eq:pstar}) of the latent representation of AEs with $n=12$ latent nodes and different depths. Each curve refers to a different detaset: 2M contains the point corresponding to digits $2$ of MNIST, 2M+ is derived from 2M adding data points obtained through translations and rotations of the original data points. M is  MNIST, EM is EMNIST and FEM includes EMNIST and Fashion MNIST. The colour code of points corresponds to the fitted value of $g$. The inset reports the values of $g$ as a function of the Kullback-Leibler divergence, for different layers rendered by the colour code (same symbols as in the main figure). {\bf c)} List of the ten most probable configurations of the latent space with their frequency for the AE with $6$ hidden layers trained on the FEM dataset.}
        \label{fig:AE}
\end{figure}

AEs are deterministic neural networks based on real valued variables. In order to express the latent representation in terms of binary variables, we adopt a sigmoid transfer function between the last layer of the encoder and the latent space, so that latent variables can be interpreted as probabilities from which we sample binary variables. 
We construct an empirical distribution $\hat p(\bs)$ by sampling ten states $\bs\in\{0,1\}^n$ from the activity patterns generated by each point in the dataset. This is then analysed with the same method described in Section~\ref{sec:gauge} in order to compute the Kullback-Leibler divergence to the HFM. 


Fig.~\ref{fig:AE} {\bf a)} shows the network architectures used in this study. These are designed so that the AE with $L+1$ hidden layers is obtained from that with $L$ layers adding a further layer between the last and the bottleneck, for both the encoder and the decoder. This procedure is meant to mimic the addition of a further coarse graining step in the information processing before the bottleneck\footnote{The data in Fig.~\ref{fig:AE} {\bf b,c)} correspond to averages over $6$ AE trained from scratch from a random initialization, for each dataset. This is different from the procedure followed for DBNs where each DBN was trained starting from the parameters estimated on the narrower dataset.}. With increasing depth and expanding breadth, we expect that the latent representation approaches $p^*$ of Eq.~(\ref{eq:pstar}). 

Fig.~\ref{fig:AE} {\bf b)} corroborates this expectation. It reports the Kullback-Leibler divergence of the latent representation with $n=12$ variables from $p^*$, 
as a function of depth $L$ for different datasets, and it shows that the latent space representation approaches $p^*$ both as depth and as breadth increase. 
This occurs, for each depth $L$, when the dataset ($\blacksquare$) containing only the digit `2' of MNIST is expanded to contain translations and rotations of the digit `2' ($\Box$), and when it is expanded to contain all the digits in MNIST ($\blacktriangledown$), then all the characters in EMNIST ($\blacktriangle$), and finally when Fashion MNIST is further added to EMNIST ($\bullet$).

It is also interesting to observe that the estimated values of $g$ approach the critical point $g_c=\log 2$ both as depth and as breadth increase (see inset of Fig.~\ref{fig:AE} {\bf b}). For $g=g_c$ the distribution $p^*$ coincides with a uniform distribution\footnote{Fitting the internal representations of AEs with the HFM $h_n$ yields values of $g$ which are in the range $[0.15,0.5]$ for $L\ge 4$. These plots show a similar behaviour to that in Fig.~\ref{fig:AE}.}. Yet the representation in the latent space is very different from a uniform distribution. Fig.~\ref{fig:AE} {\bf c)} shows that the ten most frequent states of the latent variables (after the transformations of Sect.~\ref{sec:gauge})  
for the AE with $L=6$ trained on the broadest dataset (FEM) coincide with the ten most probable states of $p^*$. This similarity holds also for other values of $L$ and datasets.

\section{Discussion}
\label{sec:end}

The main original contribution of this paper is to argue that, in layered information-processing systems such as artificial neural networks and the brain, universal representations emerge spontaneously from the combined effects of depth and breadth. We derive such universal, abstract representation as a fixed point of a RG transformation and we present numerical experiments corroborating this picture.

Abstraction is typically defined in relative terms, depending on what details are considered relevant. The process of abstraction connects different levels of detail in a hierarchy of representations. This process, we argue, can be described by an RG procedure that allows us to define {\em absolute abstraction} as the fixed point of the RG transformation, which describes the ultimate result of this process, in the limits of infinite detail or infinite generality. The distance from this limit allows us to define a quantitative measure of the level of abstraction of a representation as the Kullback-Leibler distance from the fixed point.

The fact that optimal representations of data rely on universal distributions is a well known fact in source coding. In fact, compression algorithms translate complex datasets into random strings of zeros and ones of maximal entropy~\cite{CoverThomas}. This is also true of generative models such as Variational Auto-encoders~\cite{kingmaauto} and Diffusion Models~\cite{sohl2015deep}, that map the data to vectors of  variables with a preassigned distribution. In these cases, as in source coding, information on the specific nature of the data is stored in the parameters that govern how the data is transformed. For example, in a neural network, the parameters of the different layers capture how each state of a deep representation is dressed by features at different scales, in order to generate data points. The internal representation -- the code -- just describes how the data is organized, and it is universal because it is the solution of an information theoretic optimization principle. In fact, the only common characteristic of data coming from very different domains is the coding cost, i.e. the number of bits needed to efficiently code each data point. The principle of maximal relevance~\cite{marsili2022quantifying} predicates that the coding cost should be as broadly distributed as possible, which in turn facilitates a robust alignment of different data sources along this dimension. The HFM arises as the ideal abstract representation because it satisfies the principle of maximal relevance, which qualifies it as an optimal scaffold for organizing data according to their coding cost\footnote{Note that, in the HFM, the coding cost $-\log h_n(\bs)=g m_{\bs}-\log Z_n$ depends linearly on the sufficient statistics $m_{\bs}$. So the coding cost is itself a sufficient statistics.}. 

This perspective makes the difference between fitting -- i.e. estimating the parameters that reproduce the data -- and learning -- i.e. describing the variation of the data -- clear. In addition, integrating data into a pre-existent representations makes learning more similar to understanding, and it endows it with desirable properties for intelligent behaviour\footnote{At the same time, constraining the internal representation to a data-independent, preassigned model facilitates learning~\cite{kingmaauto} and promotes interpretability~\cite{higgins2017beta}.}, as suggested in Ref.~\cite{xie2022random}. In fact, the capacity of abstraction is a key ingredient of intelligent behaviour. Abstract representations provide the map of the universe of possibilities that intelligence can navigate to "handle entirely new tasks that only share abstract commonalities with previously encountered situations."~\cite{chollet2019measure}.



The archetypal example of an abstract representation is language. In its general traits, the perspective drawn here resonates with the Chomskyan approach to linguistics, which has been very influential. According to Chomsky~\cite{chomsky2014aspects}, one has to distinguish a deep structure -- that encodes abstract semantic structures as well as grammatical rules -- and a surface structure which is derived from the deep one through a series of transformations leading to the actual, observable form of language as it is spoken or written. The deep structure entails an innate generative process -- the {\em universal grammar} -- which is argued to be common to all human languages, and which relies on the capacity of infinite recursion~\cite{hauser2002faculty} thus making it possible to generate an infinite variety of sentences with a finite vocabulary. The fact that this capacity emerges in children without exposure to much data (spoken language)~\cite{pinker2003language} has led to the hypothesis that universal grammars need to be biologically hardwired, an hypothesis that is not widely accepted~\cite{tomasello2005constructing}. It is tempting to speculate that universal grammars could emerge in deep cortical areas as fixed points of a transformation such as the one discussed here, driven by the integration of inputs from a broad set of sources, across all sensory modalities. Such universal representations would then be shaped by data which is not limited to language. In this view, it would be the integration of all experience into the same framework -- that one may call understanding -- that promotes abstraction, with the emergence of universal representations. 

Understanding how the conceptual framework developed here can be extended to more complex domains such as language is an interesting avenue of further research\footnote{In this respect, we note that the HFM can easily be generalised to variables $x_i$ taking value in an arbitrary set $\chi$ by invoking a transformation $\sigma: \chi\to \{0,1\}$ that maps each value of $x_i$ into a binary variable $s_i=\sigma(x_i)$~\cite{Betelli2024}. Extending this analysis in the time dependent domain constitutes a considerably more challenging avenue.}.
In this paper we confine ourselves to the admittedly oversimplified domain of static representations of binary variables. Also, the range of breadth of the data that our numerical experiments explore is rather limited, as well as the representation capacity of the neural networks analysed. {\color{black} Yet, these results align with empirical studies of large neural networks trained on heterogeneous datasets which exhibit a convergence toward shared representations~\cite{huh2024position}. This suggests that the RG approach discussed here may be extended to more complex data structures.}



\section{Acknowledgements} 

We are grateful to Paolo Muratore and Davide Zoccolan for interesting discussions and to Giulia Betelli for her contribution~\cite{Betelli2024}. We acknowledge Max Planck Research School (IMPRS) for The Mechanisms of Mental Function and Dysfunction (MMFD) for supporting Elias Seiffert.

\appendix

\section{Background}

In this Section we review theoretical concepts mentioned in the main paper.

\subsection{The relevance}
\label{sec:rel}

This Section recalls the definition of the relevance. We refer to Ref.~\cite{marsili2022quantifying} for a more extended treatment.
We describe representations $p(\mathbf{s})$ in terms of the variable $E_{\mathbf{s}}=-\log_2 p(\mathbf{s})$, which is the minimal number of bits needed to represent state $\mathbf{s}$. 
The average coding cost $H[\mathbf{s}]=\mathbb{E}\left[ E_{\mathbf{s}}\right]$ is the usual Shannon entropy and counts the number of bits needed to describe one point of the dataset. Following Ref.~\cite{marsili2022quantifying}, we shall call $H[\mathbf{s}]$ the {\em resolution}. 

The resolution $H[\mathbf{s}]$ is a measure of information content but not of information ``quality". Meaningful information should bear statistical signatures that allow it to be distinguished from noise. These make it possible to identify relevant information {\em before} finding out what that information is relevant for, a key feature of learning in living systems. 
Following Ref.~\cite{marsili2022quantifying}, we take the view that the hallmark of meaningful information is a broad distribution of coding costs. Here breadth can be quantified by the {\em relevance}, which is the entropy of the coding cost $E_{\mathbf{s}}$ 
%
%
%
\begin{equation}
\label{eqHE}
H[E]=-\sum_E p(E)\log_2 p(E)\,,
\end{equation}
where $p(E)=W(E)2^{-E}$ is the probability that a state $\mathbf{s}$ randomly drawn from $p(\mathbf{s})$ has $E_{\mathbf{s}}=E$, and $W(E)$ is the number of states $\mathbf{s}$ with $E_{\mathbf{s}}=E$. The {\em principle of maximal relevance}~\cite{marsili2022quantifying} postulates that maximally informative representations should achieve a maximal value of $H[E]$, which correspond to a uniform distribution of coding costs ($p(E)={\rm const}$ or $W(E)=W_0 2^{E}$).
Representations where coding costs are distributed uniformly should be promoted for the reason that, in an optimal representation, the number $W(E)$ of states $\mathbf{s}$ that require $E$ bits to be represented should match as closely as possible the number ($2^E$) of codewords that can be described by $E$ bits. Representation of maximal relevance with a given resolution also have an exponential degeneracy of states $W(E)=W_0 e^{\nu E}$, with $\nu$ that depends on $H[\mathbf{s}]$.
Note that states $\mathbf{s}$ and  $\mathbf{s}'$ with very different coding costs $E_{\mathbf{s}}$ and  $E_{\mathbf{s}'}$ can be distinguished by their statistics, because they would naturally belong to different typical sets\footnote{By the law of large numbers, typical samples of weakly interacting variables all have approximately the same coding cost, a fact knowns as the asymptotic equipartition property~\cite{CoverThomas}. 
As in Ref.~\cite{shwartz2017}, 
we take the view that a trained neural network distinguishes the points in a dataset in different typical sets.}. Representations that maximize the relevance harvest this benefit in discrimination ability that is accorded by statistics alone.

\subsection{The Hierarchical Feature Model}
\label{sec:HFM}

The HFM was introduced in~\cite{xie2024simple}, to which we refer for a more complete treatment. 
The HFM 
encodes the principle of maximal relevance. It describes the distribution of a string $\mathbf{s}_{1:n}=(s_1,\ldots,s_n)$ of binary variables that we take as indicators of whether each of $n$ features is present ($s_i=1$) or not ($s_i=0$). Features are organised in a hierarchical scale of detail and we require that the occurrence of a feature $s_k=1$ at level $k$ does not provide any information on whether lower order features are present or not. This means that, conditional on $s_k=1$, all lower order features are as random as possible, i.e. $H[\mathbf{s}_{1:k-1}|s_k=1]=k-1$ in bits. This requirement implies that the Hamiltonian $E_{\mathbf{s}}$ should be a function of $m_{\mathbf{s}}=\max\{k:~s_k=1\}$, with $m_{\mathbf{s}}=0$ if $\mathbf{s}=\mathbf{0}$
is the featureless state\footnote{This is because $p(\mathbf{s}|m_{\mathbf{s}}=k)=2^{-k+1}$ so that $p(\mathbf{s})=p(\mathbf{s}|m_{\mathbf{s}})p(m_{\mathbf{s}})=p(m_{\mathbf{s}})/2^{m_{\mathbf{s}}-1}$.} ($s_i=0~\forall i$). Since there are $2^{k-1}$ states with $m_{\mathbf{s}}=k$, the principle of maximal relevance (i.e. the requirement that $W(E)=W_0 e^{\nu E}$) excludes all functional forms between $E_{\mathbf{s}}$ and $m_{\mathbf{s}}$ that are not linear. This leads to the HFM, that assigns a probability
\begin{equation}
h_n(\mathbf{s})=\frac{1}{Z_n}e^{-g m_{\mathbf{s}}}\,,
\end{equation}
to state $\mathbf{s}$, where the partition function $Z_n$ ensures normalisation. We refer to~\cite{xie2024simple} for a detailed discussion of the properties of the HFM. In brief, in the limit $n\to\infty$ the HFM features a phase transition at $g_c=\log 2$ between a random phase where $H[\mathbf{s}]$ is of order $n$ for $g<g_c$, and a ``low temperature'' phase where $h_n(\mathbf{s})$ is dominated by a finite number of states (and $H[\mathbf{s}]$ is finite in the limit $n\to\infty$). 

Marginalising over the low order features $\mathbf{s}_{1:k}$ returns a mixture between a HFM over the remaining $n-k$ features and a frozen state 
\begin{equation}
\sum_{\mathbf{s}_{1:k}}h_n(\mathbf{s}_{1:n})=\frac{\xi^k Z_{n-k}}{Z_n}h_{n-k}(\mathbf{s}_{k+1:n}) + \frac{(1-\xi/2)(\xi^k-1)}{(\xi-1)Z_n}\delta_{\mathbf{s}_{k+1:n},\mathbf{0}_{k+1:n}}
\,.
\end{equation}
On the other hand, marginalising over the high order ones yields a mixture between the HFM and the uniform (maximum entropy) distribution
\begin{equation}
\label{eq:margtop}
\sum_{\mathbf{s}_{k+1:n}}h_n(\mathbf{s}_{1:n})=\frac{Z_k}{Z_n} h_{k}(\mathbf{s}_{1:k})+\left(1-\frac{Z_k}{Z_n} \right) 2^{-k}\,.
\end{equation}

\section{Details of analytical calculations}

In this Section we provide details for the derivations in the theoretical part of the main paper.

\subsection{Existence and uniqueness of the solution for $\alpha$}
\label{app:alpha}

Each step in the coarse graining RG involves a change in entropy $\Delta_{k\to k+1}H=H_{k+1}[s]-H_k[s]$ as follows: 
\begin{eqnarray}
\Delta_{1\to 2}H & = & H[s_{1:n-1}]-H[s_{1:n}]=-H[s_n|s_{1:n-1}] \\
\Delta_{2\to 3}H & = & H[s_0,s_{1:n-1}]-H[s_{1:n}]=1 ~~~\hbox{(bits)}\\
\Delta_{3\to 4}H & = & h(q)-\alpha H[s_{1:n}]-(1-\alpha)h(\tilde p_n(0_{1:n})),\qquad q=\alpha+(1-\alpha)\tilde p_n(0_{1:n})
\end{eqnarray}
where $h(x)=-x\log_2 x-(1-x)\log_2(1-x)$. Overall the change in entropy is
\begin{equation}
\label{ }
\Delta_{1\to 4}H=h(q)-\alpha H[s_{1:n}]-(1-\alpha)h(\tilde p_n(0_{1:n}))+1-H[s_n|s_{1:n-1}]
\end{equation}
therefore $\alpha$ is the solution of the equation $\Delta_{1\to 4}H=0$. Notice that $\Delta_{1\to 4}H(\alpha=0)\ge 0$ and 
\begin{equation}
\label{ }
\frac{d}{d\alpha}\Delta_{1\to 4}H=(1-\tilde p_n(0_{1:n}))\log\frac{1-q}{q}-H[s_n|s_{1:n-1}]+h(\tilde p_n(0_{1:n}))
\end{equation}
which is negative at $\alpha=0$ and
\begin{equation}
\label{ }
\frac{d^2}{d\alpha^2}\Delta_{1\to 4}H=-\frac{1-\tilde p_n(0_{1:n})}{q(1-q)}<0
\end{equation}
which means that the solution it is unique provided that $\Delta_{1\to 4}H(\alpha=1)=- H[\bs_{1:n}]+1-H[s_n|\bs_{1:n-1}]$ is negative. A sufficient condition is that $H[\bs_{1:n}]>1$.

\subsection{Proof of Eq.~(\ref{eq:pstar})}
\label{app:fixedpoint}

Let us first analyse how the HFM transforms under $\Re_\uparrow$. Marginalisation on $s_n$ yields
\begin{equation}
\label{ }
h_n(\bs_{1:n-1})=\sum_{s_n=0,1}h_n(\bs_{1:n})=\frac{Z_{n-1}}{Z_n}h_{n-1}(\bs_{1:n-1})+\frac{1}{Z_n}e^{-gn}
\end{equation}
Hence 
\begin{equation}
\label{ptilde}
\tilde p_n(\bs_{1:n}')=\frac{Z_{n-1}}{2Z_n}h_{n-1}(\bs_{2:n}')+\frac{1}{2Z_n}e^{-gn}
\end{equation}
If $\bs_{2:n}'=0$ then 
\[
h_{n-1}(\bs_{2:n}')=\frac{Z_{n}}{Z_{n-1}}h_n(\bs_{1:n}')e^{g s_1'}=\frac{Z_{n}}{Z_{n-1}}h_n(\bs_{1:n}')\left[e^g+(1-e^g)\delta_{s_1',0}\right]
\]
because $m_{\bs_{1:n}}=s_1$ in this case. 
If $\bs_{2:n}'\neq \mathbf{0}_{2:n}$ instead
\[
h_{n-1}(\bs_{2:n}')=\frac{Z_{n}}{Z_{n-1}}e^g h_n(\bs_{1:n}')
\]
because $m_{\bs_{2:n}}=m_{\bs_{1:n}}-1$. Therefore, both cases are accounted for by the equation
\begin{equation}
\label{ }
h_{n-1}(\bs_{2:n}')=\frac{Z_{n}}{Z_{n-1}}e^g h_n(\bs_{1:n}')+\frac{1-e^g}{Z_{n-1}}\delta_{\bs_{1:n}',0}
\end{equation}
Substituting this into Eq.~(\ref{ptilde}) yields
\begin{equation}
\label{ }
\tilde p(\bs_{1:n}')=\frac{e^g}{2}h_n(\bs_{1:n}')+\frac{1-e^g}{2 Z_n}\delta_{\bs_{1:n}',0}+\frac{e^{-gn}}{2Z_n}
\end{equation}
and 
\begin{equation}
\label{ }
p_n'(\bs_{1:n}')=(1-\alpha)\frac{e^g}{2}h_n(\bs_{1:n}')+\left[\alpha-(1-\alpha)\frac{e^g-1}{2 Z_n}\right]\delta_{\bs_{1:n}',0}+(1-\alpha)\frac{e^{-gn}}{2Z_n}
\end{equation}
Therefore, at least for finite $n$, the HFM is not a fixed point.


We look for a fixed point of the form
\begin{equation}
\label{pfix}
p_n^*(s_{1:n})=(1-\beta)h_n(s_{1:n})+\beta u_n(s_{1:n})
\end{equation}
exploiting the fact that $\Re_\uparrow$ is a linear transformation. The uniform distribution $u_n(s_{1:n})=2^{-n}$ transforms as $\Re_\uparrow(u_n)(\bs_{1:n})=(1-\alpha)2^{-n}+\alpha\delta_{\bs_{1:n},0}$. 

After some calculation, with $\xi=2 e^{-g}$,  we find
\begin{eqnarray}
p_n'(\bs_{1:n}') & = & \frac{(1-\beta)(1-\alpha)}{\xi}h_n(\bs_{1:n}') \label{l1}\\
 & ~ & +\left[(1-\beta)\left(\alpha-(1-\alpha)\frac{2-\xi}{2\xi Z_n}\right)+\beta\alpha\right]\delta_{\bs_{1:n}',0} \label{l2} \\
 & ~ & + \left[\frac{(1-\beta)(1-\alpha)}{2Z_n}{\xi^n}+\beta(1-\alpha)\right]u_n(\bs_{1:n}')\label{l3}
\end{eqnarray}
Setting the coefficient of $h_n(s_{1:n}')$ in the first line~(\ref{l1}) to $1-\beta$ and the coefficient of $u_n(s_{1:n}')$ in the third line ~(\ref{l3}) equal to $\beta$ yields
\begin{equation}
\label{pfixpar}
\alpha=1-\xi,\qquad \beta=\frac{\xi^{n+1}}{2-\xi}
\end{equation}
the second line~(\ref{l2}) then vanishes by normalization. The solution then reads
\begin{equation}
\label{eq:pss}
p^*_n(\bs_{1:n})=\left(1-\frac{1}{e^g-1}\right)e^{-gm_{\bs_{1:n}}}+\frac{1}{e^g-1}e^{-gn}\,.
\end{equation}
Interestingly, a solution only exists for $\xi<1$, i.e. for $g>g_c$, and in the limit $g\to g_c$ the fixed point distribution tends to $u_n$. 
Eq.~(\ref{eq:pss}) has the same form of an HFM with $m>n$ features, marginalised over the $m-n$ most detailed ones (see Eq.~\ref{eq:margtop}). 
The value of $m$ can be computed equating $1-\beta$ to the coefficient of $u_n$ in the marginal of $h_m(\bs_{1:m})$ over $\bs_{1:n}$. 
This yields the equation
\begin{equation}
\label{ }
\frac{\xi^{m+1}(2-\xi-\xi^{n+1})}{(2-\xi)(2-\xi-\xi^{m+1})}=0
\end{equation}
whose only solution for $\xi<1$ is $m=+\infty$. In other words, the fixed point $p_n^*$ is the marginal distribution of the $n$ most coarse grained features of an HFM with infinite features, which is Eq.~(\ref{eq:pstar}).

\section{Data and deep neural networks}
\label{app:numerics}

This Section describes first the data used in this study and then the architectures that have been trained on them.

\subsection{Data}
\label{Data DBN}
We used four standard image datasets: MNIST~\cite{lecun1998gradient}, EMNIST letters~\cite{cohen2017emnist}, Fashion-MNIST~\cite{xiao2017fashionmnistnovelimagedataset}, and CIFAR-10~\cite{Krizhevsky2009LearningML} downloaded via the torchvision library \cite{paszke2019pytorch}. 
Every dataset was binarized before training DBNs (not for auto-encoders) with the following procedure:
For MNIST, EMNIST Letters, and FMNIST (all 28×28 grayscale), we thresholded pixel intensities at 0.5 to map inputs to \{0,1\}. For CIFAR-10 (originally 32×32 RGB), each image was first converted to grayscale and then binarized with the same threshold then reduced to 28×28 via a centered crop that removes a 2-pixel border on each side.

Below we report each dataset's sizes and concise descriptions:\\
MNIST: 60 000 handwritten digits (10 classes, 28×28);\\
EMNIST Letters: 124 800 handwritten letters (26 classes, 28×28; upper/lowercase merged);\\
Fashion-MNIST: 60 000 clothing items (10 classes, 28×28);\\
CIFAR-10: 50 000 natural images (10 classes, originally 32×32 RGB). 

For sequential DBN training, we constructed datasets of increasing breadth as follows: we first partitioned MNIST into six label sets containing \{1, 2, 4, 6, 8, 10\} digit classes; we then augmented with EMNIST Letters in three steps: (i) letters a–i, (ii) letters l–r, and (iii) the remaining letters, followed by Fashion-MNIST and, finally, CIFAR-10. 

\subsection{DBNs and training procedure}
\label{app:DBN}

\begin{figure}[h!]
        \centering
        \includegraphics[width=5cm]{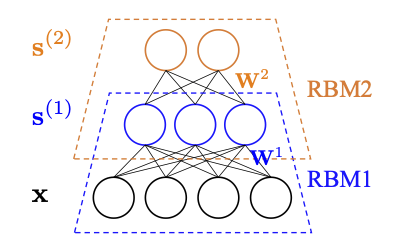}
        \caption{A three layer Deep Belief Network}
        \label{DBN}
\end{figure}

A deep belief networks (DBN) consists of Restricted Boltzmann Machines (RBM) stacked one on top of the other, as shown in Fig.\ref{DBN}. Each RBM is a Markov random field with pairwise interactions defined on a bipartite graph of two non interacting layers of variables: visible variables $\textbf{x}=(x_1,..,x_m)$ representing the data, and hidden variables $\textbf{s}=(s_1,...,s_n)$ that are the latent representation of the data. The probability distribution of a single RBM is:
\begin{equation}\label{rbm mesure}
    p(\textbf{x},\textbf{s})=\frac{1}{Z}\exp{\left(\sum_{i,j}W_{ij}x_is_j+\sum_k x_kc_k+\sum_l s_lb_l\right)}.
\end{equation}
where $\textbf{W}=\{W_{ij}\},~\textbf{c}=(c_1,\ldots,c_m)$ and $\textbf{b}=(b_1,\ldots,b_n)$ are the parameters that are learned during training.

In order to generate samples from the trained DBN we consider the connections between the top two layers as undirected, whereas all lower layers are connected to the upper layer by directed connections. This means that, in order to obtain a sample from a DBN  we initialized a random configuration $\hat{\mathbf{s}}^{\ell}_0$, and performed alternating Gibbs sampling for $10^6$ steps to ensure convergence to the equilibrium distribution of the top RBM $p_{L}(\textbf{s}^{(L)},\textbf{s}^{(L-1)})$. Then we use this data to sample the states of lower layers using the conditional distribution $p(\mathbf{s}_{\ell-1}|\mathbf{s}_{\ell})$. In this way, we propagate the signal till the visible layer. 

The DBN used in our experiment is the same as that used in Ref.~\cite{Song}: it has a visible layer with $784$ nodes and $L=10$ hidden layers with the following number of nodes: $n_\ell=500, 250, 120, 60, 30, 25, 20, 15, 10$ and $5$, for $\ell=1,\ldots,10$.

We train the DBN one layer at a time, following Hinton's prescription \cite{hinton2006fast}. First, the bottom RBM is fitted to the data; the dataset $\hat{\mathbf{x}}=(\mathbf{x}_1,\ldots,\mathbf{x}_N)$ is then propagated to the first hidden layer to obtain samples $\hat{\mathbf{s}}^{(1)}$, which become the training set for the next RBM. This type of training procedure was proven \cite{hinton2006fast} to increase a variational lower bound for the log-likelihood.

For each RBM, parameters are learned by stochastic gradient ascent on the log-likelihood, using Persistent Contrastive Divergence with $k=10$ (PCD-10), a learning rate of $0.01$, and mini-batches of size $100$ (see~\cite{hinton2012practical}), for $\sim 10^5$ epochs. With these settings the training of the DBNs does not exhibit mode collapse, which typically occurs when the dataset is strongly clustered and the persistent Markov chain fails to properly mix across all clusters.

We trained the DBN sequentially on datasets of progressively increasing breadth (see Data \ref{Data DBN}). At each stage, we restarted training with the same settings (PCD-10, learning rate, mini-batch size, number of epochs), initializing all weights with those learned at the previous stage; we then trained the entire DBN on the extended dataset.


Decelle et al. \cite{decelle2021equilibrium} \cite{agoritsas2023explaining} showed that an RBM trained with CD-10 does not reproduce the equilibrium distribution, yet it can still serve as a good generative model when sampled out of equilibrium. Instead, persistent contrastive divergence (PCD-10), which we use, tends to learn a better equilibrium distribution.\footnote{In Contrastive 
Divergence-k (CD-k), the Markov chain used to sample the distribution is initialized on the batch used to compute the gradient and $k$ Monte Carlo steps are performed. In Persistent Contrastive Divergence-k (PCD-k) the MCMC is initialized in the configuration of the previous epoch.}




\subsection{Autoencoder Architecture and Training Procedure}
\label{app:AE}

The autoencoders employed in this study were trained on the MNIST dataset and its variants \cite{lecun1998mnist}. Consequently, all models described below accept an input vector of dimension \(784\).

\subsubsection{Architecture}

Each model implements a fully connected, feed-forward autoencoder composed of two principal components: an \emph{encoder} that maps the high-dimensional input into a compact latent representation, and a \emph{decoder} that attempts to reconstruct the original input from that representation. The encoder and decoder are constructed symmetrically around a central \emph{bottleneck} (latent) layer, so that the decoder mirrors the encoder’s layer sizes and activation choices in reverse  (see Fig.~\ref{fig:AE} {\bf a}). This symmetric design facilitates interpretation of the learned mapping between input space and latent space and is a common architectural choice for classical autoencoders \cite{hinton2006reducing}.

The number of hidden layers explored across experiments ranges from one to seven. Let \(n_0=784\) denote the input dimension and let \(n_i\) denote the number of neurons in the \(i\)-th hidden layer of the encoder. Successive encoder layers are defined by a fixed geometric reduction factor of \(0.6\); specifically,
\[
n_{i+1} = \big\lceil 0.6\, n_i \big\rceil,
\]
where \(\lceil\cdot\rceil\) denotes the ceiling function to ensure that layer sizes are integer-valued. This progressive compression produces a sequence of hidden sizes that decreases smoothly from the input dimension down to the pre-specified bottleneck dimensionality and provides a controlled way to vary model capacity and depth while keeping the reduction schedule consistent across models.

\begin{figure}[t!]
        \centering
        \includegraphics[width=15cm]{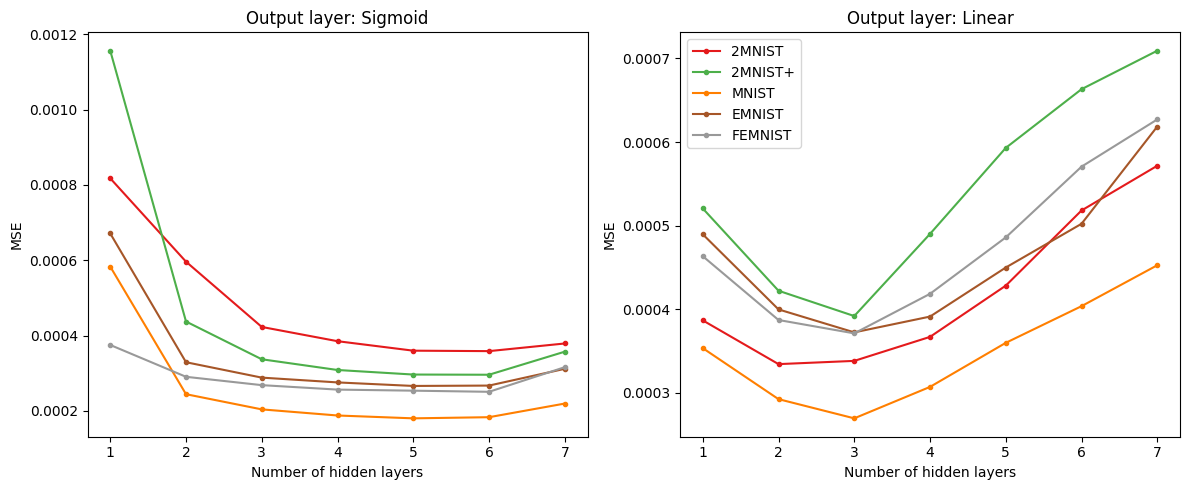}
        \caption{The Mean Squared Error of the autoencoders networks with ten latent dimensions, for the case of sigmoid as output layer (left) and the case of a linear layer (right). In the latter case, the increase of the error for networks with more than three hidden layers coincides with the preference of the network for memorizing data points, rather than detecting features.} 
        \label{fig:AElosses}
\end{figure}

Activation functions were assigned as follows: all intermediate hidden layers employ the rectified linear unit (ReLU) nonlinearity, defined as \(\mathrm{ReLU}(x)=\max(0,x)\); ReLU units are widely used owing to their favorable optimization properties (reduced saturation and mitigated vanishing gradients) \cite{nair2010rectified}. Both the layer immediately preceding the bottleneck and the last layer of the decoder use a sigmoid activation \(\sigma(x) = (1+e^{-x})^{-1}\). Attempts of using as output layer the linear function were made, but the networks showed a preference for memorizing datapoints instead of detecting features. Moreover, the Mean Squared Error (see \ref{Training procedure}) is sensibly lower and better behaved in the case of the sigmoid function as output (see Fig. \ref{fig:AElosses}). The latent (bottleneck) dimension itself was treated as a variable: distinct networks were trained with latent dimensions $d=10$ and $12$ 
in order to study the effect of representational capacity on reconstruction quality and on the geometry of learned codes.

The decoder is symmetric to the encoder: starting from the latent layer, successive layers increase dimensionality according to the inverse schedule (mirroring the reduction factor of the encoder) until the reconstructed output of dimension \(784\) is produced.

A symmetric architecture reduces the number of arbitrary design choices and often yields decoder features that are naturally aligned with encoder features, easing interpretability of the coding/decoding mapping. In classical autoencoders one can choose to tie decoder weights to the transpose of encoder weights (reducing parameter count and imposing linear symmetry) or allow untied weights (more flexibility at the cost of more parameters). The present work uses untied weights, which provides additional decoder capacity and avoids forcing a linear transpose constraint; the choice was motivated by the aim of assessing how representational dimensionality (bottleneck size) and depth affect reconstruction independent of a tied-weight regularization.

\subsubsection{Training procedure}\label{Training procedure}

All networks were trained using the Adam optimizer with learning rate \(\eta=10^{-3}\) and \(L_2\) weight decay (also referred to as \emph{weight decay}) set to \(10^{-5}\). Adam is an adaptive first-order optimizer that computes individual learning rates for each parameter from estimates of first and second moments of the gradients and is well suited for stochastic optimization problems that are nonstationary and noisy \cite{kingma2014adam}. The training objective used for all runs was the mean squared error between the input \(x\) and its reconstruction \(\hat{x}\):
\[
\mathcal{L}_{\mathrm{MSE}}(x,\hat{x}) \;=\; \frac{1}{D}\sum_{j=1}^{D}\big(x_j - \hat{x}_j\big)^2,
\]
where \(D=784\) is the input dimensionality. MSE is a natural choice when the goal is to minimize Euclidean reconstruction error; for image data preprocessed to lie in \([0,1]\), MSE provides a direct measure of pixelwise fidelity.
Each architecture/latent dimension configuration was trained for 10 epochs through the training set.

\paragraph{Layer-wise pretraining and initialization strategy}

To improve optimization stability and to provide favorable initializations for deeper models, a greedy layer-wise pretraining procedure was employed. The pretraining approach used here follows the pragmatic spirit of early deep representation learning strategies in which layers are learned progressively and representations learned by shallower architectures are used to initialize deeper ones \cite{hinton2006reducing,bengio2007greedy}... The precise procedure implemented is as follows:

\begin{enumerate}
  \item Train a shallow autoencoder with a small number of hidden layers (for instance, one hidden layer plus a bottleneck). Training at this stage optimizes the MSE objective for the shallow architecture starting from standard random initialization.
  \item When the shallow model has converged, retain the learned weights for the encoder part up to the deepest layer of that shallow model.
  \item Construct a deeper model with one additional hidden layer. Initialize the weights of the new deeper model as follows: (i) copy the encoder weights from the previously trained shallow model into the corresponding positions in the deeper model; (ii) initialize the newly added layers (both encoder and matching decoder layers) with random weights (e.g., small Gaussian i.i.d.\ initialization); (iii) mirror the copied encoder initializations into the corresponding decoder positions if using symmetric initialization heuristics.
  \item Train the deeper model on the same reconstruction objective until convergence.
  \item Repeat the expansion and initialization procedure iteratively until the target depth (up to seven hidden layers in this study) is reached.
\end{enumerate}

This greedy scheme yields an initialization for each deeper architecture from parameters that have already learned useful lower-level features; empirically, such initialization can reduce the likelihood of becoming trapped in poor local minima and can accelerate convergence relative to training the deepest architecture from random initialization \cite{hinton2006reducing,bengio2007greedy}..

\paragraph{Regularization and normalization}

Batch normalization and dropout were intentionally omitted in order to maintain full transparency of the learned latent representations and to avoid introducing additional mechanisms that could obscure the relationship between architecture depth, bottleneck dimensionality, and reconstruction characteristics. Batch normalization can significantly alter the distribution of activations during training and often improves optimization speed and stability \cite{ioffe2015batch} ; dropout randomly zeros activations during training and acts as an implicit model ensemble/regularizer \cite{srivastava2014dropout}. While both are valuable tools for improving generalization in many supervised and unsupervised contexts, their use would complicate a direct, controlled investigation of how depth and bottleneck size alone influence the autoencoder’s representational geometry.

\bibliographystyle{unsrt}
\bibliography{abstraction}

\end{document}